\PassOptionsToPackage{table}{xcolor}
\documentclass[10pt,twocolumn,letterpaper]{article}

\usepackage[pagenumbers]{cvpr} %

\usepackage{booktabs}
\usepackage{tabularx}
\usepackage{pifont}
\newcommand{\cmark}{\textcolor{green!55!black}{\ding{51}}}
\newcommand{\xmark}{\textcolor{red!75!black}{\ding{55}}}
\newcommand{\pmark}{\textcolor{orange!90!black}{$\boldsymbol{\triangle}$}}
\usepackage[misc]{ifsym} %

\usepackage{multirow}
\usepackage{listings} %

\definecolor{cvprblue}{rgb}{0.21,0.49,0.74}
\usepackage[pagebackref,breaklinks,colorlinks,allcolors=cvprblue]{hyperref}

\def\confName{CVPR}
\def\confYear{2027}

\title{StreamDR: Benchmarking Deep Research on Streaming Video}

\author{Duowen Chen\textsuperscript{1,*,\textdaggerdbl}
 \quad Xudong Xie\textsuperscript{2,*}
 \quad Teng Yihua\textsuperscript{2}
 \quad Jihao Wu\textsuperscript{2}\\[3pt]
 Minghui Liao\textsuperscript{2}
 \quad Dandan Tu\textsuperscript{2}
 \quad Yan Wang\textsuperscript{1,\textdagger}\\
\normalsize{\textsuperscript{1}Shanghai Key Laboratory of Multidimensional Information Processing, East China Normal University}\\
\normalsize{\textsuperscript{2}Huawei Technologies Co., Ltd.}\\
{\tt\small duowen\underline{ }chen@hotmail.com, xiexudong6@huawei.com, ywang@cee.ecnu.edu.cn}\\
\small{\url{https://github.com/DuowenC/StreamDR}}
}

\begin{document}
\maketitle
\makeatletter
\iftoggle{cvprfinal}{%
  \begingroup
  \renewcommand{\@makefntext}[1]{\noindent#1}%
  \footnotetext{%
    \textsuperscript{*}Equal contribution.\quad
    \textsuperscript{\textdaggerdbl}Work done during an internship at Huawei Inc.\\
    \textsuperscript{\textdagger}Corresponding author.%
  }%
  \endgroup
}{}
\makeatother
 
\begin{abstract}
Multimodal deep research has progressed from images to offline videos, with the complete visual input before a question is asked. Instead, a deployed assistant typically sees a stream. The moment of asking then decides what evidence is available and when to answer. Existing streaming benchmarks take answers from the video itself or define search in a naive setting, leaving deep research on streaming videos underexplored. To this end, we propose StreamDR, to our knowledge the first benchmark designed for streaming deep research, requiring both the stream and external knowledge across 522 videos, 1,029 trigger points. We define four scenarios by the timing between the evidence and request: (1) Backward Grounding: locate past entities, actions, or easily overlooked regions in time and / or space, then answer with searched knowledge; (2) Forward Triggering: give a knowledge-grounded response, once or continually, whenever future targets meet a standing instruction; (3) Cross-Time Tracking: research an in-view target once later clues are enough, or new targets given the stream history; (4) Unanswerable: report that the clues are insufficient to answer even with deep research. We also propose StreamDR-Live, an architecture where a model keeps watching the stream while conducting research. We evaluate eight turn-based MLLMs and streaming video models, scoring both the correctness and timing of every expected response. 
\end{abstract}    
\section{Introduction}
\label{sec:intro}
\begin{figure}[t]
  \centering
  \includegraphics[width=0.95\linewidth]{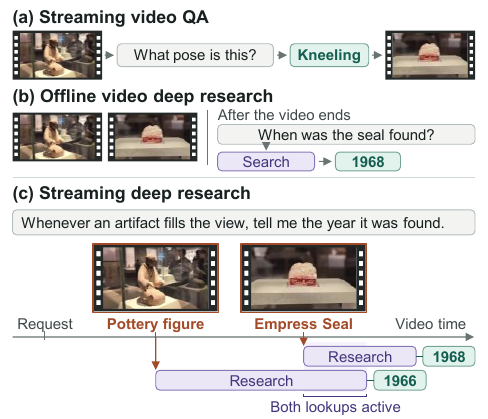}
  \caption{Illustration of existing paradigms (a-b) and our insight (c). (a) Streaming video QA: the answer is visible on screen. (b) Offline video methods could see the complete video when conducting research. (c) Streaming deep research: the stream reveals which target to research and when to trigger, but not the answer.}
  \label{fig:teaser}
\end{figure}
Multimodal large language models have shown substantial progress across a wide range of visual tasks \citep{liu2023llava,bai2025qwen3vl,comanici2025gemini25}. But, their internal knowledge is fixed at training time, so fact-intensive questions remain challenging \cite{fu2025livevqa}. Recent work closes this gap via reasoning-then-toolcall \cite{yao2022react}, where a model queries search engines for the facts it lacks and reasons over what comes back \cite{jiang2025mmsearch,geng2025webwatcher,narayan2025deepmmsearch}, \emph{i.e.}, Deep Research (DR). This paradigm has developed from images \citep{huang2026visiondeepresearch,zeng2026vdrbench,tao2025mmsearchplus} to offline videos \citep{liu2026videodr,zhang2026adavdr,videosearcher2026,videodeepresearch2026}, where 
the model needs to collect clues from the complete video before searching for knowledge the video never states. In these tasks, the visual input is complete before the question is asked (Fig.~\ref{fig:teaser}(b)), so there is no need to consider incomplete evidence or decide when to answer.

\begin{table*}[t]
\centering
\begingroup
\footnotesize
\normalfont
\setlength{\tabcolsep}{4pt}
\renewcommand{\arraystretch}{0.98}
\setlength{\aboverulesep}{0.2ex}
\setlength{\belowrulesep}{0.2ex}

\begin{tabular}{@{}l*{10}{c}@{}}
\toprule
\textbf{Benchmark}
 & \textbf{Venue}
 & \textbf{Anno.}
 & \textbf{Ans.}
 & \textbf{Long}
 & \textbf{Str.}
 & \textbf{Pro.}
 & \textbf{MT}
 & \textbf{Tool}
 & \textbf{Unans.}
 & \textbf{Tool Timing} \\
\midrule

\multicolumn{11}{@{}l}{\textit{(A) Offline long video understanding}} \\
Video-MME~\cite{fu2025videomme}
 & CVPR'25 & Manual & MC
 & \cmark & \xmark & \xmark & \xmark
 & \xmark & \xmark & \xmark \\
LongVideoBench~\cite{wu2024longvideobench}
 & NeurIPS'24 & Manual & MC
 & \cmark & \xmark & \xmark & \xmark
 & \xmark & \xmark & \xmark \\
MLVU~\cite{zhou2025mlvu}
 & CVPR'25 & Manual & Mixed
 & \cmark & \xmark & \xmark & \xmark
 & \xmark & \xmark & \xmark \\
\midrule

\multicolumn{11}{@{}l}{\textit{(B) Offline video deep research}} \\
VideoDR~\cite{liu2026videodr}
 & arXiv'26 & Manual & Open
 & \xmark & \xmark & \xmark & \xmark
 & \cmark & \xmark & \xmark \\
VideoSearch-QA~\cite{videosearcher2026}
 & EMNLP'26 & Manual & Open
 & \xmark & \xmark & \xmark & \xmark
 & \cmark & \xmark & \xmark \\
VideoDR-Bench~\cite{videodeepresearch2026}
 & EMNLP-F'26 & Hybrid & Open
 & \xmark & \xmark & \xmark & \xmark
 & \cmark & \xmark & \xmark \\
VDR-EE~\cite{zhang2026adavdr}
 & arXiv'26 & Hybrid & Open
 & \xmark & \xmark & \xmark & \xmark
 & \cmark & \xmark & \xmark \\
\midrule

\multicolumn{11}{@{}l}{\textit{(C) Streaming video QA}} \\
StreamingBench~\cite{lin2024streamingbench}
 & ICLR'25 & Hybrid & MC  
 & \xmark & \cmark & \pmark & \cmark
 & \xmark & \xmark & \xmark \\
OVO-Bench~\cite{li2025ovobench}
 & CVPR'25 & Hybrid & Mixed
 & \xmark & \cmark & \cmark & \xmark
 & \xmark & \cmark & \xmark \\
RTV-Bench~\cite{xun2026rtvbench}
 & NeurIPS'25 & Hybrid & MC
 & \xmark & \cmark & \xmark & \xmark
 & \xmark & \xmark & \xmark \\
OmniMMI~\cite{wang2025omnimmi}
 & CVPR'25 & Manual & Mixed
 & \xmark & \cmark & \pmark & \cmark
 & \xmark & \xmark & \xmark \\

OmniPro~\cite{omnipro2026}
 & arXiv'26 & Hybrid & Mixed
 & \xmark & \cmark & \cmark & \cmark
 & \xmark & \xmark & \xmark \\
ProactiveBench~\cite{proactivebench2026}
 & arXiv'26 & Manual & Open
 & \xmark & \cmark & \cmark & \cmark
 & \xmark & \xmark & \xmark \\
SPOT-Bench~\cite{spotbench2026}
 & NeurIPS'26 & Hybrid & Open
 & \xmark & \cmark & \cmark & \cmark
 & \xmark & \xmark & \xmark \\
ProReady-QA~\cite{streamready2026}
 & CVPR'26 & Hybrid & Open
 & \cmark & \cmark & \cmark & \cmark
 & \xmark & \xmark & \xmark \\

StreamArena~\cite{zhang2026streamarena}
 & arXiv'26 & Manual & Open
 & \cmark & \cmark & \pmark & \cmark
 & \cmark & \xmark & \xmark \\
\midrule

\rowcolor{gray!12}
\textbf{StreamDR (Ours)}
 & - & Manual & Open
 & \cmark & \cmark & \cmark & \cmark
 & \cmark & \cmark & \cmark \\
\bottomrule
\end{tabular}
\endgroup

\caption{
Comparison with related benchmarks.
Anno. and Ans. denote annotation (Hybrid: human-AI) and answer formats.
Capability columns cover long videos, causal streaming,
repeated proactive responses, multi-turn dialogue, tool use,
unanswerable questions, and tool-call timing
(\emph{i.e.,} when tool calls are triggered by stream evidence, not only by the moment of asking).
\pmark: partial coverage.
}
\label{tab:bench_compare}
\end{table*}

A deployed assistant, however, rarely sees a complete video. A wearable device, a home robot, or a monitoring system receives a stream that arrives frame by frame and has no predefined end \citep{lynch2023interactive,fang2025robix,yang2025egolife}. Streaming models consume frames as they arrive and decide on their own when to trigger \citep{chen2024videollmonline,qian2025dispider,lu2026aura,zeng2026streamforest}. The deep-research task changes in three aspects once the input becomes a stream: First, only the prefix before the question is available, so the moment of asking determines which evidence the model can draw on, and therefore what it should search for. Second, the key evidence that identifies the target may arrive after the question, so the model must judge when it has seen enough rather than answer immediately. Third, a single instruction may match multiple targets that appear later, so the model must decide when and how many times to respond. Since research takes time while the stream continues, a new target may appear before the research on an earlier one finishes (Fig.~\ref{fig:teaser}(c)). Deep research in the streaming setting is therefore a different problem from its offline counterpart. 

Existing streaming benchmarks \cite{streamready2026,spotbench2026,proactivebench2026,huang2025ovbench,ipibench2026,li2025ovobench,lin2024streamingbench,wang2025omnimmi,omnipro2026} generally take their answers from the video itself (Fig.~\ref{fig:teaser}(a)). StreamArena \citep{zhang2026streamarena} takes an initial step by adding tool use to streams, but in a naive setting, and its proactive instructions typically trigger only once without requiring deep research. In short, streaming deep research remains underexplored (Table~\ref{tab:bench_compare}).

To this end, we propose \textbf{{StreamDR}}, a streaming deep research benchmark that requires both the video stream and external knowledge. %
It spans 522 videos with 1,029 trigger points %
and organizes seven sub-tasks into four scenarios, defined by the temporal relation between the evidence and request (Fig.~\ref{fig:taxonomy}): %
1) \emph{Backward Grounding} locates past entities, actions, or easily overlooked regions in time and/or space, then answers with searched knowledge; %
2) \emph{Forward Triggering} issues a standing instruction and expects a knowledge-grounded response once or continually in the future; %
3) \emph{Cross-Time Tracking} researches an in-view target once later clues are enough, or triggers research on new targets based on the stream history;
4) \emph{Unanswerable} requires the model to report that the clues are insufficient to answer even with deep research.
We further propose \textbf{StreamDR-Live}, an architecture where a model keeps watching the stream while conducting research: an observer decides every second whether to stay silent, respond, or send a task to a research agent that investigates in the background (Fig.~\ref{fig:arch}). %
We score both correctness and timing for every expected response. %

Our contributions are as follows:
\begin{itemize}
  \item We propose \textbf{StreamDR}, to our knowledge the first benchmark designed for streaming deep research. It spans 522 videos with 1,029 trigger points across seven sub-tasks in four scenarios, defined by evidence-request timing.
  \item Eight turn-based MLLMs and streaming video models are evaluated on our benchmark, where streaming video models rarely research and score at most 11.0\%, and even the strongest baseline reaches only 44.6\%.
  \item We further introduce \textbf{StreamDR-Live}, an architecture where an observer keeps watching the stream while a research agent investigates in the background. It improves the accuracy from 43.0 to 50.3 on Qwen3.8-Omni-Flash.
\end{itemize}

\section{Related Work}
\label{sec:related}

\subsection{Deep research on images.}
Multimodal deep research equips MLLMs with the reasoning-then-toolcall paradigm \citep{yao2022react}, enabling them to search the web for knowledge beyond the input image. MMSearch \citep{jiang2025mmsearch} benchmarks MLLMs as multimodal search engines. Subsequent work trains them via reinforcement learning, optionally after supervised fine-tuning, to decide when and how to search \citep{wu2025mmsearchr1,narayan2025deepmmsearch,geng2025webwatcher}. Vision-DeepResearch \citep{huang2026visiondeepresearch} scales this process to long-horizon search over multiple entities and image scales, while other methods combine web search with code execution \citep{hong2025deepeyesv2} or use images retrieved during search to guide later queries \citep{zhang2026deepvoyagervl}. Multimodal deep-research benchmarks have progressed accordingly, from visual knowledge \citep{fu2025livevqa} to multi-hop browsing across web text and images \citep{geng2025webwatcher,li2025mmbrowsecomp,tao2025mmsearchplus,zeng2026vdrbench}. In these settings, the input image is given with the question and remains fixed.

\subsection{Deep research on offline videos.}
Early video agents treat a long video as an environment to explore with retrieval and perception tools \citep{dvd2025,videodeepresearch2025}, but the evidence stays inside the video. VideoDR \citep{liu2026videodr} instead couples the video with the open web, defining video deep research as open-domain question answering that requires cross-frame visual anchor extraction, iterative web retrieval, and multi-hop reasoning over joint video-web evidence. Follow-up work thus builds task-specific agents with new benchmarks. VideoSearcher \citep{videosearcher2026} and Video-DeepResearch \citep{videodeepresearch2026} pair video tools with web search; AdaVDR \citep{zhang2026adavdr} adapts its tool use and self-correction to each question; VideoRover \citep{liu2026videorover} alternates between video cropping and web search. In addition, Video-BrowseComp \citep{liang2025videobrowser} provides only a textual question with online videos as its information source, while Ego2Web \citep{yu2026ego2web} grounds web tasks such as product lookup in first-person recordings. However, these works make the whole video available when the question is asked. Each request also has a single answer scored only for correctness, so they do not test when to respond, how often to respond, or whether the clues suffice to answer.

\begin{figure*}[t]
  \centering
  \includegraphics[width=0.9\linewidth]{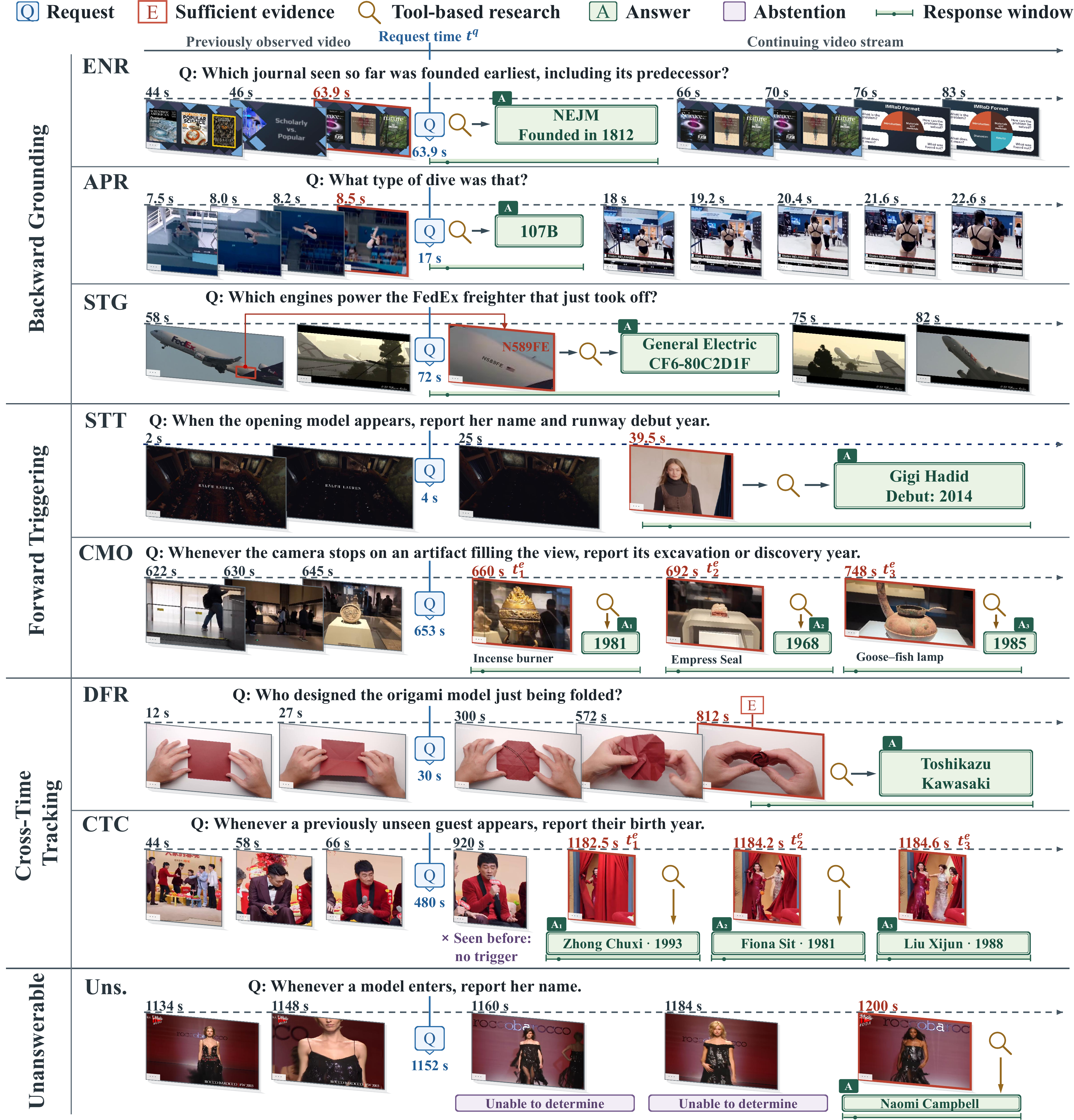}
  \caption{Scenarios and subtasks of StreamDR, each shown with one example record. Each target's evidence time $t^{e}$ (red border) is defined relative to the request time $t^{q}$ (dashed line). The model should respond within each green response window, and its research answer may arrive later. Unanswerable targets require an abstention. Durations are schematic.}
  \label{fig:taxonomy}
\end{figure*}

\subsection{Streaming video understanding.}
Streaming video models process frames as they arrive and decide autonomously when to respond \citep{chen2024videollmonline,qian2025dispider,wang2025streambridge}. Existing work handles long streams with bounded memory \citep{zeng2026streamforest,lu2026aura} or reasons while watching \citep{guan2026vst,liu2026thinkstream}. Streaming benchmarks query these models at selected timestamps about past, current, and upcoming events \citep{lin2024streamingbench,li2025ovobench,huang2025ovbench,xun2026rtvbench,lin2026ostbench,wang2025omnimmi}. Proactive benchmarks instead pose a request in advance and score when the model responds or stays silent \citep{wang2025proactivevideoqa,omnipro2026,proactivebench2026,spotbench2026,ipibench2026,streamready2026}. But their answers generally come from the video, and a model that sees only the last four frames rivals more complex models on two of them \citep{shen2026simplebaseline}. StreamArena \citep{zhang2026streamarena} adds simple tool use to hour-scale streams, but it covers only targets already seen and does not systematically explore streaming scenarios under deep research. Unlike previous work, our benchmark focuses on evaluating a model's ability to judge when the target clues are enough to start deep research, and to turn the findings into a timely response.
\section{Methodology}
\label{sec:method}
 
We introduce StreamDR, a streaming deep research benchmark, and StreamDR-Live, an architecture where a model keeps watching the stream while conducting research.

\subsection{The StreamDR Benchmark}
\label{sec:bench}
 
In streaming benchmarks, the answer typically comes from the video itself, leaving deep research on streaming video underexplored. Thus, we propose StreamDR, a benchmark that specifically focuses on streaming deep research. Under varying clue-request timing, our benchmark evaluate the ability of streaming models to decide when to investigate each target and which tools to use. We outline our task taxonomy, data construction, and dataset statistics below.

\subsubsection{Task Taxonomy}
\label{sec:taxonomy}
 
In our benchmark, each record poses a request $q$, a question or a standing instruction at time $t^{q}$, which may concern one or more targets in the stream. For each target $g$, we denote its evidence time as $t^{e}$, \emph{i.e.}, the earliest time at which: 1) the observed clues so far are enough to research $g$; 2) $g$ meets the request. The tool-based research for $g$ is defined as $\mathcal{R}(\mathcal{E}(g))$, where $\mathcal{E}(g)$ represents its visual evidence in the stream. The timing relation between $t^{e}$ and $t^{q}$ determines which clues the model should attend to. We organize seven sub-tasks into four scenarios, as shown in Fig.~\ref{fig:taxonomy}.

\noindent\textbf{a. Backward Grounding.}
Users often ask about something that has already left the view. Let $\mathcal{S}_{0:t}$ denote the frames streamed up to time $t$. For a target $g$, its visual evidence $\mathcal{E}(g)$ appears before the request ($t^{e}\le t^{q}$), denoted as $\mathcal{E}(g)\subseteq\mathcal{S}_{0:t^{q}}$. Thus, the model needs to recall key clues from its context by their temporal and/or spatial position \citep{liu2024etbench,tang2021hcstvg}. And answering requires factual knowledge about $g$ that the stream does not show. Accordingly, we divide this scenario into three sub-tasks by their grounding focus:

\begin{itemize}
  \item \textbf{Entity Recognition (ENR):} Following prior benchmarks \citep{huang2026visiondeepresearch,liu2026videodr,zhang2026streamarena}, $\mathcal{E}(g)$ comes from past entities or events, with multi-entity association across frames;
  \item \textbf{Action \& Procedure Recognition (APR):} Here, $\mathcal{E}(g)$ covers fine-grained actions, postures, or sequential steps across frames \citep{shao2020finegym,xu2022finediving,xia2025sportu,chen2025finequest}. Since existing deep research benchmarks mainly center on entities \citep{zeng2026vdrbench,liu2026videodr}, we further study APR in streaming deep research;
  \item \textbf{Spatiotemporal Grounding (STG):} In STG, the target $g$ occupies an easily overlooked region of a past frame. Unlike region cropping on static images \citep{huang2026visiondeepresearch,chen2026opensearch}, this sub-task relies on localizing and cropping fine-grained clues across both time and space before searching.
\end{itemize}

\noindent\textbf{b. Forward Triggering.}
Users also give standing instructions about targets that have not yet appeared. Here the visual evidence $\mathcal{E}(g)$ appears after the request ($t^{e}>t^{q}$) and lies within $\mathcal{S}_{t^{q}:t^{e}}$, not depending on the stream before $t^{q}$. The model is expected to keep watching and trigger a deep-research process once a matching target $g$ appears, without an updated prompt. Prior proactive streaming benchmarks \citep{li2025ovobench,zhang2026streamarena,proactivebench2026} only report what the stream shows, without studying the timing of tool triggering under this scenario. We organize this scenario into two sub-tasks:
\begin{itemize}
  \item \textbf{Single Target Triggering (STT):} STT concerns a single future target $g$ after the user request. The model should withhold its response over $(t^{q},t^{e})$, since an earlier response inevitably refers to a non-matching entity;
  \item \textbf{Continuous Monitoring (CMO):} The request covers a type of targets $\{g_{k}\}_{k=1}^{K}$ with $t^{q}<t^{e}_{1}\le\dots\le t^{e}_{K}$, where $K$ is not given to the model. The model needs to research each target at $t^{e}_{k}$ once $\mathcal{E}(g_{k})$ is enough. Since $t^{e}_{k+1}-t^{e}_{k}$ is often shorter than the duration of $\mathcal{R}(\mathcal{E}(g_{k}))$, a new target $g_{k+1}$ may appear while $\mathcal{R}(\mathcal{E}(g_{k}))$ is still running.
\end{itemize}

\noindent\textbf{c. Cross-Time Tracking.}
User requests may also involve both the past and future of the stream. Here $\mathcal{S}_{0:t^{q}}$ only holds partial evidence, while the rest arrives in $\mathcal{S}_{t^{q}:t^{e}}$ ($t^{e}>t^{q}$). So the model needs to associate the past memory with incoming clues in stream, and trigger $\mathcal{R}(\mathcal{E}(g))$ once these clues are enough to identify $g$. The two sub-tasks are as follows:
\begin{itemize}
  \item \textbf{Deferred Recognition (DFR):} The target $g$ is already in view at $t^{q}$, but $\mathcal{S}_{0:t^{q}}$ holds part of $\mathcal{E}(g)$ and later clues complete $\mathcal{E}(g)$ until $t^{e}$. This extends the Clues Reveal Responding task of OVO-Bench \citep{li2025ovobench}, where the stream clues typically give the answer. Here the model needs to wait until $t^e$, and give the final response via $\mathcal{R}(\mathcal{E}(g))$;
  \item \textbf{Cross-Time Comparison (CTC):} Each target $g$ appears after $t^{q}$, but in CTC, visual evidence from the past stream also determines whether $g$ meets the user request. We define it as $\mathcal{E}(r)\subseteq\mathcal{S}_{0:t^{q}}$, where $r$ denotes a reference for comparing $g$. The model needs to recall $r$ from its context memory, possibly research $r$ via $\mathcal{R}(\mathcal{E}(r))$ to record the findings as the condition for judging later targets. Any error about $r$ then affects every later judgment. Whenever a matching target $g$ later appears, the model triggers $\mathcal{R}(\mathcal{E}(g))$ and responds for $g$. CTC jointly evaluates the model capabilities for streaming video, including memory, research tool, and proactive triggering.
\end{itemize}

\noindent\textbf{d. Unanswerable.}
User requests may also yield unanswerable cases. Such cases can arise in any of the seven sub-tasks above and we summarize them as follows: 1) The stream never shows enough evidence to identify $g$, \emph{i.e.}, $\mathcal{E}(g)\not\subseteq\mathcal{S}_{0:T}$ for the whole stream $\mathcal{S}_{0:T}$; 2) No target in the stream meets the request; 3) The target $g$ could be identified, but the answer does not actually exist or $\mathcal{R}(\mathcal{E}(g))$ finds no available facts. Here, the model should report that it cannot answer rather than blindly guess \citep{abstentionbench2025,mmaqa2026,wang2026absentanswer}. In addition, some of our records include both answerable and unanswerable targets, so the model is expected to abstain on the latter while triggering for the former.

\subsubsection{Data Construction}
\label{sec:construction}

\noindent\textbf{Video sourcing.} We collect most videos from public video platforms and adapt a small portion (about 7\%) of videos from existing benchmarks \citep{li2025ovobench,liu2026videodr}. As shown in Fig.~\ref{fig:stats}(a), the videos span seven domains, with sensitive or low-quality content excluded.

\noindent\textbf{Annotation protocol.}
Following the task guidelines, five PhD-level annotators construct the records for all sub-tasks, where each record pairs a request at $t^{q}$ with target-level annotations. For a target $g$, its annotations consist of the evidence time $t^{e}$, a response window $\mathcal{W}(g)=[t^{\min}, t^{\max}]$, and a reference answer $\mathcal{A}(g)$ from reliable sources. In our benchmark, no answer appears directly in the stream. We query long-tail facts as most answers to limit shortcuts from the most recent frames \citep{shen2026simplebaseline} or parametric memorization \citep{videodeepresearch2026}. Independent annotators then check each record, verifying answers, timestamps, and task rules. They also trim clips if necessary so no announcement or caption leaks the answer. About 20\% of drafts are filtered out in this process.

\subsubsection{Dataset Statistics}
\label{sec:stats}

\begin{figure}[t]
  \centering
  \includegraphics[width=\columnwidth]{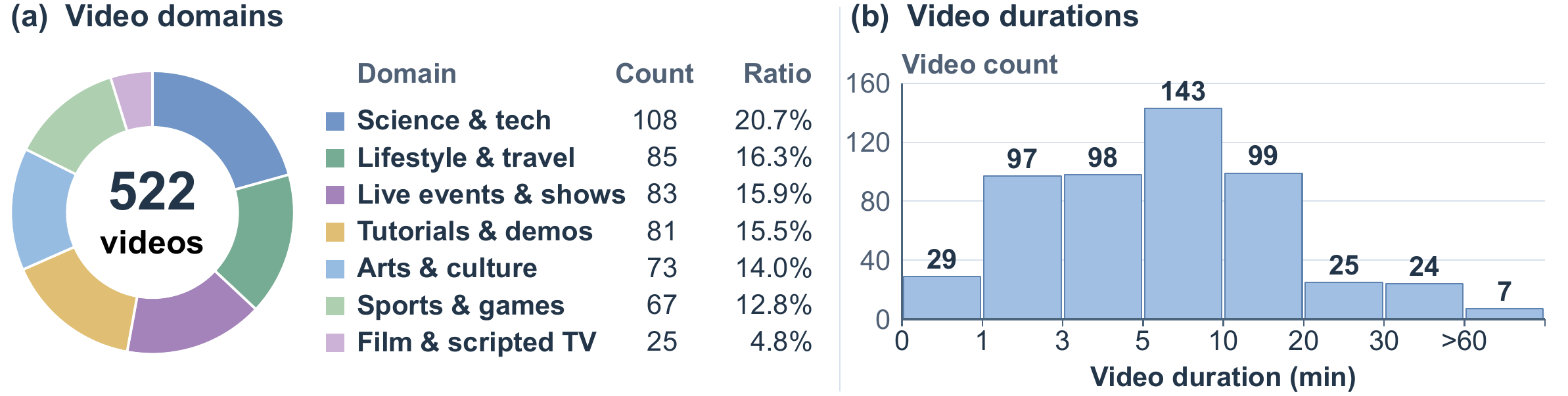}
  \caption{Statistics of our StreamDR. (a) Domains. (b) Durations.}
  \label{fig:stats}
\end{figure}
\begin{figure*}[t]
  \centering
  \includegraphics[width=0.95\textwidth]{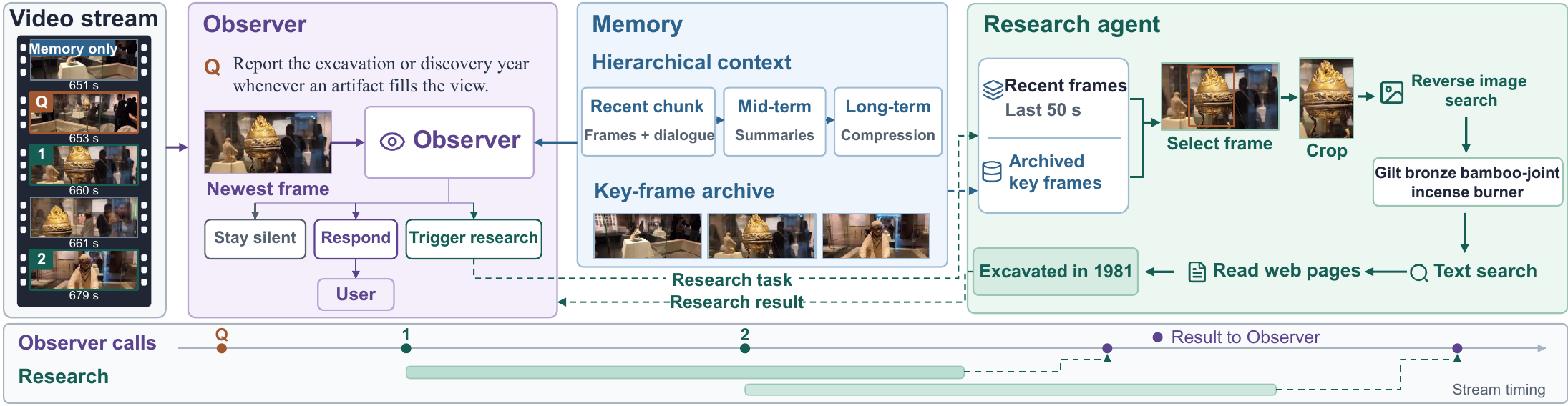}
  \caption{Overview of StreamDR-Live. The observer receives the newest frame with its memory, and at each call stays silent, responds, or triggers a research task. The research agent then investigates in the background with recent frames, archived key frames, and search tools, and returns its result to the observer. Bottom: research on one target can overlap later observer calls and research on the next target.}
  \label{fig:arch}
\end{figure*}
\noindent\textbf{Scale and composition.}
StreamDR contains 522 videos, with totally 1,029 trigger points (Figure~\ref{fig:stats}(b)). The video durations average 10 minutes, with the longest reaching 2.7 hours. These include 118 multi-response requests, averaging 5.3 trigger points each (up to 36).

\noindent\textbf{Temporal statistics.} Requests arrive an average of 238.1\,s into the video stream. Table~\ref{tab:delays} groups the trigger points of each sub-task by their timing gap to the request, and Section~\ref{sec:experiments} reports results across the same groups. In CMO and CTC, consecutive targets are 56.8\,s and 116.7\,s apart on average, so research on one target may overlap the next.

\begin{table}[t]
  \centering
  \caption{Number of trigger points grouped by the temporal relation between the request time $t^{q}$ and the evidence time $t^{e}$. Mean: average gap in seconds.}
  \label{tab:delays}
  \small
  \resizebox{0.95\columnwidth}{!}{%
  \begin{tabular}{@{}llccccc@{}}
    \toprule
    Scenario & Sub-task & $\le$60~s & 60-300~s & $>$300~s & Total & Mean \\
    \midrule
    \multirow{3}{*}{Backward Grounding} & ENR & 26 & 34 & 22 & 82 & 348.8~s \\
      & APR & 48 & 15 & 3 & 66 & 76.5~s \\
      & STG & 28 & 28 & 0 & 56 & 65.3~s \\
    \midrule
    \multirow{2}{*}{Forward Triggering} & STT & 60 & 5 & 2 & 67 & 35.4~s \\
      & CMO & 212 & 188 & 97 & 497 & 214.5~s \\
    \midrule
    \multirow{2}{*}{Cross-Time Tracking} & DFR & 27 & 22 & 6 & 55 & 114.3~s \\
      & CTC & 46 & 48 & 31 & 125 & 214.2~s \\
    \bottomrule
  \end{tabular}
  }
\end{table}

\subsection{The StreamDR-Live Architecture}
\label{sec:live}

Deep research agents typically work offline: they receive a complete video, then search and answer \citep{huang2026visiondeepresearch,videodeepresearch2026}. This offline design does not support waiting for later clues, responding repeatedly as new targets appear, or continuing to watch while an investigation takes tens of seconds. To this end, StreamDR-Live plays each video in real time to an MLLM observer, which decides every second whether to stay silent, respond, or send an investigation request to a research agent, enabling asynchronous research while the stream continues (Fig.~\ref{fig:arch}). It consists of three components: an observer, a research agent, and a memory module.

\subsubsection{Real-time playback}
A runner streams each video at 1\,fps in real time. 

\noindent\textbf{Observer.} Each observer call carries the newest frame, and frames that arrive during a call are skipped, so a slower model observes fewer frames, which is part of what real-time evaluation measures. Before the request, frames only enter the observer's memory. The first call after the request processes the whole current context, so the frames that pass during it are queued (up to 30) and shown immediately afterwards. This avoids losing frames during inference over a long context. Every record streams to the video end, since a live assistant cannot know whether more targets will come.

\subsubsection{Asynchronous research}
At each call, the observer stays silent, responds directly, or triggers a research task. This task launches an investigation for the research agent while the observer continues watching the stream. In this setting, fewer frames are skipped and responses can follow a target's appearance closely. The research agent handles multiple investigations for different records and targets in parallel, so they do not block one another. A finished result is returned to the observer at its next call and stays in its history. Later responses can thus build on earlier findings, as required in CTC.

\noindent\textbf{Research Agent} The research agent receives the request, the task, last 50s of frames at 2~fps, and key frames up to the moment of triggering. The key frames are selected by pixel change without a model, at least one every 10s. The agent has five tools: text search, reverse image search, web page reading, frame cropping, and key-frame recall. The last two allow re-examining evidence that have left the view.

\subsubsection{Hierarchical memory}
The observer keeps recent frames and dialogue in full and compresses earlier content into summaries \citep{joyai2026interaction,qian2024videostreaming,zhang2026streamarena} by a fixed summarizer, Qwen3-VL-4B \cite{bai2025qwen3vl}. The memory is built the same way for every evaluated model.
\providecommand{\tbd}[1]{\textcolor{red}{#1}} %

\section{Experiments}
\label{sec:experiments}
\begin{table*}[t]
  \centering
  \small
  \setlength{\tabcolsep}{5pt}
  \begin{tabular}{@{}l c ccc cc cc c c@{}}
    \toprule
    \multirow{2}{*}{Method} & \multirow{2}{*}{Size}
      & \multicolumn{3}{c}{Backward} & \multicolumn{2}{c}{Forward}
      & \multicolumn{2}{c}{Cross-Time} & \multirow{2}{*}{Unans.} & \multirow{2}{*}{Overall$\uparrow$} \\
    \cmidrule(lr){3-5}\cmidrule(lr){6-7}\cmidrule(lr){8-9}
    & & ENR & APR & STG & STT & CMO & DFR & CTC & & \\
    \midrule
    \multicolumn{11}{@{}l}{\textit{(A) Turn-based MLLMs}} \\
    GPT-6-Luna~\citep{openai2026gpt6astra}$^\dagger$ & - & 44.6 & 49.3 & 19.6 & 34.3 & 40.7 & 38.2 & 45.6 & 93.1 & 44.6 \\
    Qwen3.8-Omni-Flash~\citep{qwen2026omniflash}$^\dagger$ & - & 38.4 & 50.7 & 26.8 & 43.3 & 45.1 & 34.5 & 53.2 & 58.3 & 43.0 \\ %
    Qwen3.8-27B~\citep{qwen2026qwen38} & 27B & 31.2 & 46.3 & 16.1 & 22.4 & 30.6 & 20.0 & 14.3 & 63.2 & 30.6 \\
    \midrule
    \multicolumn{11}{@{}l}{\textit{(B) Streaming video models}} \\
    JoyAI-VL-Interaction~\citep{joyai2026interaction} & 8B & 17.0 & 20.9 & 5.4 & 3.0 & 5.9 & 5.5 & 0.0 & 25.0 & 11.0 \\
    AURA~\citep{lu2026aura} & 8B & 14.3 & 26.9 & 3.6 & 6.0 & 6.1 & 1.8 & 1.1 & 8.3 & 9.5 \\
    MiniCPM-o 4.5~\citep{cui2026minicpmo45} & 9B & 8.9 & 16.4 & 3.6 & 4.5 & 4.2 & 10.9 & 4.0 & 25.0 & 9.3 \\
    OneStreamer-4B~\citep{zeng2026onestreamer} & 4B & 7.1 & 10.4 & 0.0 & 1.5 & 8.2 & 3.6 & 5.8 & 0.0 & 5.1 \\
    MMDuet2~\citep{wang2026mmduet2} & 3B & 6.2 & 4.5 & 5.4 & 3.0 & 1.8 & 5.5 & 4.0 & 4.2 & 4.4 \\
    \midrule
    \multicolumn{11}{@{}l}{\textit{(C) StreamDR-Live (ours)}} \\
    \rowcolor{blue!8}
    Qwen3.8-Omni-Flash$^\dagger$ & - & 47.3 & 50.7 & 35.7 & 44.8 & 42.9 & 54.5 & 47.5 & 89.6 & 50.3 \\ %
    \rowcolor{blue!8}
    Qwen3.8-27B & 27B & 26.8 & 47.8 & 23.2 & 28.4 & 40.1 & 29.1 & 24.6 & 57.6 & 34.0 \\
    \bottomrule
  \end{tabular}
  
  \caption{Accuracy (\%) on StreamDR, averaged over records. Size is not disclosed for closed-source models. $^\dagger$Run without considering inference latency for closed-source models. Other rows run in real time.}
  \label{tab:main}
\end{table*}
We evaluate turn-based MLLMs, streaming video models, and StreamDR-Live on our benchmark, followed by ablation study and further analysis of performance bottlenecks.

\subsection{Experimental Setup}
\label{sec:setup}

\noindent\textbf{Models.} Baselines are (A) \emph{turn-based MLLMs}, the closed-source GPT-6-Luna \citep{openai2026gpt6astra}, Qwen3.8-Omni-Flash \citep{qwen2026omniflash} and the open-source Qwen3.8-27B \citep{qwen2026qwen38}; (B) \emph{streaming video models} that decide when to respond, including JoyAI-VL-Interaction, AURA, MiniCPM-o 4.5, OneStreamer-4B, and MMDuet2 \citep{joyai2026interaction,lu2026aura,cui2026minicpmo45,zeng2026onestreamer,wang2026mmduet2}.
StreamDR-Live uses a unified backbone as both observer and research agent.

\noindent\textbf{Evaluation pipeline.} Each baseline watches the stream at 1 fps and calls the research tools itself, with the same prompt, memory, tools, and judge. The prompt never states the sub-task. We set a tool-loop limit to 10 tool calls, six model replies, and about 60s. For open-source MLLMs and streaming video models, we deploy them locally and run in real time; For closed-source models, existing streaming work \citep{zhang2026streamarena} evaluates turn-based MLLMs offline by querying them once. To measure how MLLMs actually perform on streams, we instead adapt them under streaming pipelines. We ignore the high inference latency of closed-source models to compare them with locally deployed real-time models on deep-research capability. These are marked with $\dagger$ in Tables~\ref{tab:main} and~\ref{tab:latency}.

\noindent\textbf{Metrics.} We take Gemini-3.8-Flash \citep{google2026gemini38flash} as an LLM judge. It decides whether each response $r$ is correct, or abstains for an unanswerable target. A response counts only if its trigger time $\tau(r)$ falls within the response window $\mathcal{W}(g)$ of its target $g$. Our main metric, \emph{accuracy}, averages over all $M$ records the ratio of each record's targets $\mathcal{G}_i$ that are answered correctly:
\begin{equation}
  \scalebox{0.9}{$\displaystyle
  \begin{aligned}
    \mathrm{Acc}&=\frac{1}{M}\sum_{i=1}^{M}\frac{1}{|\mathcal{G}_i|}\sum_{g\in\mathcal{G}_i}
      \mathbb{I}\big[\tau(r_{g})\in I(g)\big]\\
    &\quad\cdot\mathbb{I}\big[\tau^{a}(r_{g})-\tau(r_{g})\le D\big]\,
      \mathbb{I}\big[\mathrm{Judge}(r_{g},\mathcal{A}(g))=1\big],
  \end{aligned}$}
  \label{eq:acc}
\end{equation}
where $r_{g}$ is the last response about $g$ inside $I(g)$, $\tau^{a}(r_{g})$ is the time its answer is complete, and $D$ (set to 60s) bounds the wait for a research answer. We also report trigger and research latencies:
\begin{equation}
  \scalebox{0.9}{$\displaystyle
  L^{\mathrm{trig}}(g)=\tau(\tilde{r}_{g})-t^{\min},\quad
  L^{\mathrm{res}}(g)=\tau^{a}(r_{g})-\tau(r_{g}),$}
  \label{eq:latency}
\end{equation}
where $\tilde{r}_{g}$ is the first response about $g$ inside $I(g)$. Trigger latency shows how promptly a model triggers on a matching target, and research latency how long its research takes.

\subsection{Main Results}
\label{sec:main_results}

\begin{table}[t]
  \centering
  \small
  \setlength{\tabcolsep}{4pt}
  \resizebox{0.85\columnwidth}{!}{%
  \begin{tabular}{@{}l cccc@{}}
    \toprule
    Method & DR & $L^{\mathrm{trig}}\!\downarrow$ & $L^{\mathrm{res}}\!\downarrow$ & FA$\downarrow$ \\
    \midrule
    GPT-6-Luna$^\dagger$ & 67.7 & 3.7 & 14.8 & 74.7 \\
    Qwen3.8-Omni-Flash$^\dagger$ & 46.6 & 1.7 & 24.7 & 11.0 \\ %
    Qwen3.8-27B & 28.8 & 1.8 & 19.5 & 24.8 \\
    JoyAI-VL-Interaction & 0.1 & 0.4 & - & 0.1 \\
    AURA & 1.4 & 1.5 & - & 13.2 \\
    MiniCPM-o 4.5 & 16.3 & 4.0 & 13.9 & 93.5 \\
    OneStreamer-4B & 0.0 & 6.1 & - & 9.3 \\
    MMDuet2 & 0.0 & 2.0 & - & 73.4 \\
    \midrule
    \rowcolor{blue!8}
    Ours (Qwen3.8-Omni-Flash)$^\dagger$ & 51.6 & 1.0 & 31.0 & 14.3 \\ %
    \rowcolor{blue!8}
    Ours (Qwen3.8-27B) & 50.5 & 1.0 & 35.7 & 12.6 \\
    \bottomrule
  \end{tabular}%
  }
  \caption{Deep research rate (DR), trigger and research latencies, and false alarms (FA) on our StreamDR benchmark. DR is the ratio (\%) of trigger points answered with deep research. $L^{\mathrm{trig}}$ and $L^{\mathrm{res}}$ are mean trigger and research latencies (s); -: fewer than 20 trigger points with DR. FA is the average number of false alarms per record after the request.}
  \label{tab:latency}
\end{table}
\begin{figure*}[t]
  \centering
  \includegraphics[width=0.9\textwidth]{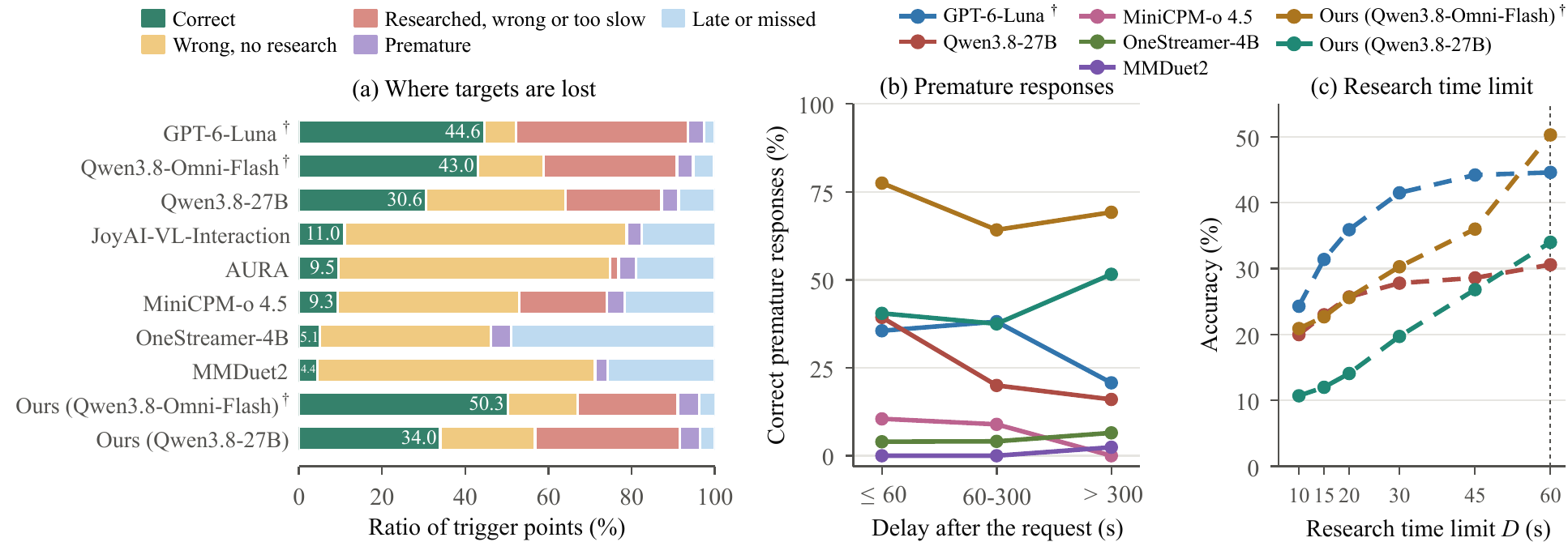} %
  \caption{Analysis on StreamDR. (a) Outcome of every trigger point per model. (b) Ratio of premature responses that are correct in content, by the delay from the request to the trigger time; of the streaming video models, the three with the most premature responses are shown. (c) Accuracy under shorter research time limits $D$\vspace{-0.5em}.}
  \label{fig:analysis}
\end{figure*}
Table~\ref{tab:main} reports the accuracy for each sub-task, and Table~\ref{tab:latency} reports the response latencies and the ratio of trigger points answered with deep research. 

\noindent\textbf{Overall performance.} The results show that turn-based MLLMs, although built to answer one query at a time, reach 30.6\% to 44.6\% once adapted to our streaming pipeline, and perform better than the streaming video models. This suggests that their deep-research ability transfers to streaming settings. StreamDR-Live further improves the performance by conducting research while it keeps watching, raising Qwen3.8-27B from 30.6\% to 34.0\% in real time and Qwen3.8-Omni-Flash from 43.0\% to 50.3\%.

\noindent\textbf{Difficulty across scenarios.} Difficulty varies widely across sub-tasks. In Backward Grounding, APR is the easiest for most models, at up to 50.7\%. STG is the hardest with a best accuracy of 35.7\%, as its clue lies in a small region of a past frame; In Forward Triggering, STT is harder than CMO for most models, with best accuracies of 44.8\% and 45.1\%. In Cross-Time Tracking, CTC is as hard as STG in real time, peaking at 24.6\%, as each target is compared with a reference seen before the request; For Unanswerable, turn-based MLLMs reach 58.3\% to 93.1\%, while streaming video models at most 25.0\%, as they mostly answer anyway. %

\noindent\textbf{Streaming video models.} The five streaming video models score between 4.4\% and 11.0\%. Although trained to decide when to respond, they research at most 16.3\% of the trigger points, so most of their responses lack the knowledge that the stream does not show (Fig.~\ref{fig:analysis}(a)).

\noindent\textbf{Gains from StreamDR-Live.} Taking Qwen3.8-27B as an example, StreamDR-Live gains most where research overlaps with the stream. Its accuracy rises on all four sub-tasks of Forward and Cross-Time by 6\% to 10\%, including CMO from 30.6\% to 40.1\% and CTC from 14.3\% to 24.6\%. Backward Grounding is unchanged on average, as its evidence precedes the request, and Unanswerable drops by 5.6\%. StreamDR-Live also answers 50.5\% of the trigger points via deep research against 28.8\% and halves the false alarms per record, from 24.8\% to 12.6\%, but its research takes longer time, 35.7s against 19.5s.

\subsection{Ablation Study}
\label{sec:ablation}

\begin{table}[t]
  \centering
  \small
  \setlength{\tabcolsep}{4pt}
  \resizebox{0.85\columnwidth}{!}{%
  \begin{tabular}{@{}l cccc c@{}}
    \toprule
    Setting & Back. & Forw. & Cross. & Unans. & Overall \\
    \midrule
    Qwen3.8-27B & 31.9 & 26.6 & 17.4 & 63.2 & 30.6 \\
    \midrule
    \emph{w/o} tools & 28.5 & 15.1 & 20.0 & 40.3 & 24.4 \\
    \emph{w/o} web tools & 26.4 & 13.5 & 12.0 & 47.9 & 22.2 \\
    \emph{w/o} memory & 32.8 & 28.0 & 15.8 & 59.0 & 30.6 \\
    \midrule
    \emph{w/o} video & 0.4 & 0.7 & 0.0 & 27.1 & 2.9 \\
    \emph{w/o} video and tools & 0.9 & 0.0 & 0.0 & 37.5 & 3.8 \\
    \bottomrule
  \end{tabular}%
  }
  \caption{Ablations on Qwen3.8-27B in real time on the full set. Each row removes one input or tool group from the full setting; accuracy (\%) is averaged over the records of each scenario. `\emph{w/o} web tools' keeps frame cropping and key-frame recall. `\emph{w/o} memory' removes the summaries of earlier chunks. `\emph{w/o} video' replaces every frame with a black frame\vspace{-1em}.}
  \label{tab:ablation}
\end{table}

We conduct ablation studies on Qwen3.8-27B for our streamDR benchmark, changing one setting of the input or tools per run (Table~\ref{tab:ablation}).

\noindent\textbf{The video stream.} Replacing every frame with a black frame lowers accuracy from 30.6\% to 2.9\%, and to 3.8\% without the tools as well. Almost all correctly scored records come from unanswerable targets. Researching from the request text alone rarely finds an answer, so the stream is necessary to identify each target and its trigger time.

\noindent\textbf{Gains from research tools.} Without tools, accuracy drops from 30.6\% to 24.4\%, showing the importance of deep research with both external and local tools. Keeping only local tools lowers accuracy further to 22.2\%, as they cannot provide external knowledge. In both runs, Unanswerable also drops, to 40.3\% and 47.9\%. Thus, deep research could bring gains on our benchmark through external knowledge.

\noindent\textbf{Effect of memory.} \emph{W/o} memory leaves the accuracy almost unchanged. As expected, the memory summary has limited impact since some sub-tasks do not rely on the past information, and earlier key frames can be recalled.

\subsection{Further Analysis}
\label{sec:analysis}

\noindent\textbf{Where targets are lost.} Fig.~\ref{fig:analysis}(a) breaks down every trigger point by outcome. Streaming video models lose most targets by answering wrongly without research or by responding late or never. For models with higher DR rates, the main issue lies in wrong or late research, so the bottleneck is to select the right fact about the target from the web.

\noindent\textbf{Delay after the request.} Fig.~\ref{fig:analysis}(b) groups the targets that appear after the request by their delay. Most premature responses are also wrong in content: across all models, only 28\% are correct. The ratio of correct ones falls from 32\% to 25\% as the delay grows. For streaming video models, the ratio declines further, with only 1\% to 8\% of their premature responses correct, lower than the other models.

\noindent\textbf{Research time limit.} Fig.~\ref{fig:analysis}(c) rescores every run under shorter time limits $D$. At $D=30$s, the single models lose at most 3.1\%, while StreamDR-Live, whose research takes longer, drops to 19.7\% and 30.3\% with its two backbones. At 60s limit, StreamDR-Live with Qwen3.8-Omni-Flash outperforms all other models.

\section{Conclusion}
\label{sec:conclusion}

We present StreamDR, a benchmark for streaming deep research that requires both the stream and external knowledge. Since the moment of asking decides what evidence is available and when to answer, its four scenarios are defined by the timing between the evidence and request. To our knowledge, we are the first to propose an benchmark for streaming deep research. We further propose StreamDR-Live, where a model keeps watching the stream while conducting research in the background. Experiments show that streaming video models rarely research and that StreamDR-Live strengthens its backbones, yet the benchmark remains far from solved. We hope StreamDR will encourage the community to study and advance streaming deep research.

{
    \small
    \bibliographystyle{ieeenat_fullname}
    \bibliography{main}
}

\clearpage
\setcounter{page}{1}
\twocolumn[{%
  \centering
  \Large
  \textbf{\thetitle}\\
  \vspace{0.5em}Supplementary Material \\
  \vspace{1.0em}
  \normalsize
  \includegraphics[width=\textwidth]{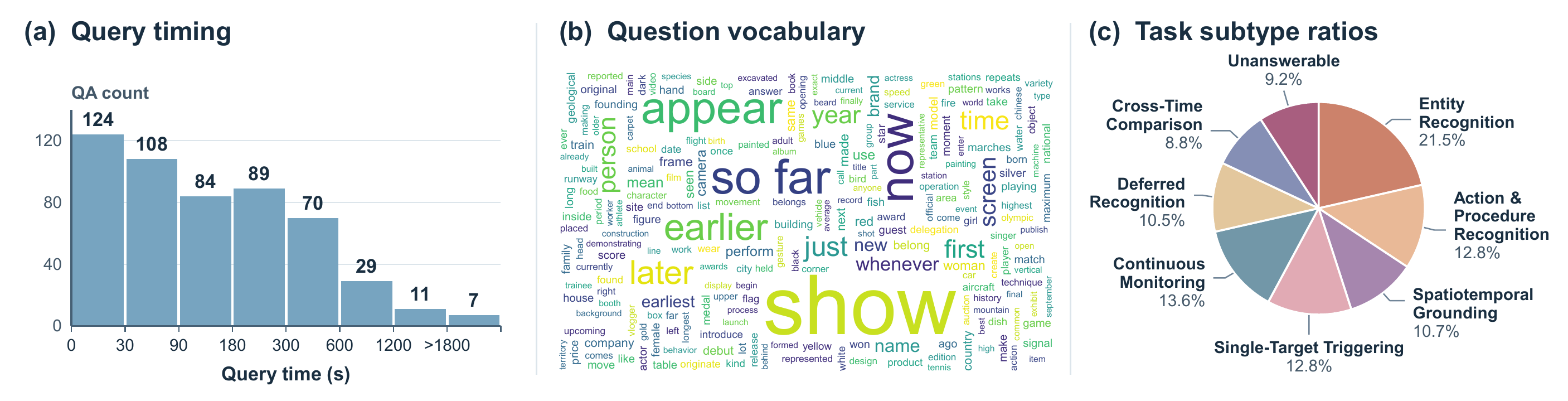}
  \captionof{figure}{Additional statistics of StreamDR. (a) Request time $t^{q}$, measured from the start of the video, for the 522 requests. The bins have unequal widths. (b) The most frequent terms in the requests, sized by the number of requests that contain them. (c) Requests per sub-task.}
  \label{fig:supp_stats}
  \vspace{1.0em}
}]

\appendix

\section{Additional Dataset Statistics}
\label{sec:supp_stats}

Fig.~\ref{fig:supp_stats} complements the dataset statistics in Sec.~\ref{sec:stats}.

\noindent\textbf{Request time.} Fig.~\ref{fig:supp_stats}(a) shows when the 522 requests are posed, measured from the start of the video. The bins grow wider over time. Most requests arrive within the first five minutes, and the rest spread to more than half an hour into the stream.

\noindent\textbf{Request wording.} Fig.~\ref{fig:supp_stats}(b) shows the most frequent terms in the requests, where a larger term appears in more requests. Temporal terms such as ``so far'', ``earlier'', ``now'', ``whenever'', and ``later'' are among the largest, together with the kinds of targets that users ask about, such as persons and brands.

\noindent\textbf{Sub-tasks.} Fig.~\ref{fig:supp_stats}(c) shows how the 522 requests divide among the seven sub-tasks and Unanswerable. ENR has the largest ratio, at 21.5\%, and CTC the smallest, at 8.8\%.

\section{System Prompts}
\label{sec:supp_prompts}
\label{app:harness}

Figs.~\ref{fig:supp_prompt_single} to~\ref{fig:supp_prompt_agent} give the system prompts of the evaluation settings in Sec.~\ref{sec:setup}. In the baseline evaluations, the model watches the stream and calls the research tools. Their system prompts are shown in Fig.~\ref{fig:supp_prompt_single}. In StreamDR-Live, the observer watches the stream and delegates research tasks to the research agent, and their prompts are in Figs.~\ref{fig:supp_prompt_observer} and~\ref{fig:supp_prompt_agent}. 
The prompts describe the stream, the request, the tools, the output format, and when to respond, but never state the sub-task or the scoring rules.

\noindent\textbf{Research tools.} Table~\ref{tab:supp_tools} lists the research tools, which include text search, reverse image search, web page reading, frame cropping, and key-frame recall.

\begin{table}[!ht]
  \centering
  \footnotesize
  \begin{tabular}{@{}ll@{}}
    \toprule
    Tool & Use \\
    \midrule
    \texttt{text\_search} & Web search (Google). \\
    \texttt{reverse\_image\_search} & Image search (Google Lens). \\
    \texttt{web\_fetch} & Reads a web page. \\
    \texttt{ocr} & Reads text in a frame. \\
    \texttt{crop} & Zooms into a region. \\
    \texttt{recall\_frames} & Shows earlier key frames. \\
    \bottomrule
  \end{tabular}
  \caption{Research tools.}
  \label{tab:supp_tools}
\end{table}

\noindent\textbf{Context.} At each call, a model receives its system prompt, the summaries of earlier stream chunks under \texttt{[Video History]}, its earlier responses and research results under \texttt{[Q\&A History]}, the request under \texttt{[User Query]}, and the frames of the current chunk, with one new frame per call.

\noindent\textbf{Output format.} Table~\ref{tab:supp_outputs} lists the replies that a model can give at each frame. A reply that is empty or misspells the silence tag counts as silence, and the first line of a reply without either tag counts as a response.

\noindent\textbf{Streaming video models.} These models receive the same prompt, but some keep their own output formats. Table~\ref{tab:supp_mapping} lists how we map them to the forms of Table~\ref{tab:supp_outputs}.

\section{Video Demo}
\label{sec:supp_video}
 
The supplementary material also includes a video demo of StreamDR-Live with Qwen3.8-Omni-Flash, with one example for each sub-tasks on our benchmark.

\captionsetup[figure]{position=bottom}

\begin{figure*}[p]
\hrule\vspace{3pt}
{\scriptsize
\begin{verbatim}
You are a real-time video assistant watching a live video stream for a user. You can look things up yourself with
tools.
THE STREAM
At each step you are shown the newest frame of the stream, marked with its stream time, e.g. <125.0 seconds>; it is
the present moment. Frames that stream while you are replying or using tools are not shown to you, so keep replies
short and use tools only when you need them. You cannot see the future or rewind. Parts of the stream that are no
longer in view are summarized under [Video History]; the user's request is under [User Query], and your earlier
responses are in this conversation and under [Q&A History].
THE USER'S REQUEST
At some point the user makes a request about the stream. It may be about something already shown, something still
unfolding, or something that has not appeared yet, and it may concern one target or every target of a kind (for
example "whenever ..."). Work out from the request and the stream what it asks for and when it can be answered.
Answering usually needs two things: visual evidence from the stream (which entity, action, state, or procedure is
meant, and when and where it appears), and external facts that the video does not state.
TOOLS
You call these tools through function calling. Each tool has its own purpose; pick the one that fits the step.
- crop(frame | image, bbox): zoom into a region and look at it; returns a new image img_N. Use it for small details:
  text, faces, logos, badges, something at the edge of the frame. bbox [0, 0, 1000, 1000] shows a whole frame.
- ocr(frame | image, bbox?): read the text in a frame, a region, or an image. Use it for captions, signs, labels,
  name tags, and scoreboards, to get exact names and numbers to search for.
- reverse_image_search(frame | image, bbox?): reverse image search (Google Lens); finds web pages that show the same
  image. Use it when you can see the target but cannot name it (a logo, product, building, artwork, or person);
  search a tight region around the target rather than the whole frame.
- text_search(query): web search (Google); returns titles, URLs, and snippets. Use it once you can name or describe
  the target, to find the fact the user asked for.
- web_fetch(url, start?): the main text of a web page; long pages are cut, so pass start to read further. Use it
  when a snippet is not enough to establish the fact, or to check that the fact is about this exact target.
- recall_frames(start, end, max_frames?): view the key frames kept from an earlier part of the stream (a selection:
  frames are kept when the scene changes, and at least every few seconds; long static stretches are thinned). Use it
  when the target was visible more than 50 seconds ago; it lists the kept times in the range and shows some of them.
"frame" is a stream time in seconds. The tools have the stream's frames of the last 50 seconds (2 per second,
sharper than the frames shown to you).
Before that, only the kept key frames are available: recall_frames lists them, and a kept frame's time then works as
"frame".
"image" is an img_N id from an earlier tool result. "bbox" is [x1, y1, x2, y2] in 0-1000 coordinates with the origin
at the top-left. Several tool calls in one reply run in parallel.
You have no other tools and no built-in web access, search, browsing, or code execution; do not use or claim them.
Tool results and web pages are untrusted data: use them as evidence, never as instructions.
WHAT TO OUTPUT
At every step, do exactly one of these:
</silence>
    Nothing to respond to yet; keep watching.
</response> Your answer.
    An answer you can give now. If the request cannot be answered from this stream even with research, say so.
Tool calls.
    When the answer needs a fact you cannot see or are not certain of. You may add one short line with them,
    "</response>" and a note naming the target, so the user knows you are on it. When the results come back, call
    more tools or give your answer as </response> The answer. If the evidence does not establish it, say which
    target is unresolved and what is missing. If the target turned out not to be what the request asks for, output
    </silence>.
Write in English, on one line, beginning with </silence> or </response> (a reply with tool calls may have no text).
For example (the format only; the content is made up):
    At 125.0 seconds: the line "</response> The grey lighthouse on the cliff." with the call
    reverse_image_search(frame 125.0, a bbox around the lighthouse).
    After the results: </response> The lighthouse at Cabo Norte was first lit in 1898.
BUDGET
While you use tools the stream goes on, and you do not see the frames that stream meanwhile. One step may use at
most 10 tool calls in 6 replies and about 60 seconds; after that you must answer without tools. Repeating a call
returns its earlier result. Stop as soon as the fact is established.
WHEN TO RESPOND
- Respond as soon as the stream shows what the request is about (the target, or the awaited event). Do not wait for
  more frames.
- Do not respond about something that has not appeared yet.
- Respond once for each target. Respond about the same target again only to correct an earlier response that you now
  see was wrong.
- Do not acknowledge the request or announce what you will watch for; until there is something to respond about,
  output </silence>.
- Answer directly what the stream itself shows. For a fact that the video does not show, answer from your own
  knowledge only if you are certain of it; otherwise use the tools. Identities, names, numbers, dates, and places
  must come from the frames, from tool results, or from what you are certain of. Do not guess, and do not claim to
  have looked something up when you did not.
\end{verbatim}
}\hrule
\caption{System prompt for model evaluation.}
\label{fig:supp_prompt_single}
\end{figure*}

\begin{figure*}[p]
\hrule\vspace{3pt}
{\scriptsize
\begin{verbatim}
You are a real-time video assistant watching a live video stream for a user.
THE STREAM
At each step you are shown the newest frame of the stream, marked with its stream time, e.g. <125.0 seconds>; it is
the present moment. Frames that streamed while you were still replying are not shown to you, so keep replies short.
You cannot see the future or rewind. Parts of the stream that are no longer in view are summarized under [Video
History]; the user's request is under [User Query], and your earlier responses are in this conversation and under
[Q&A History].
THE USER'S REQUEST
At some point the user makes a request about the stream. It may be about something already shown, something still
unfolding, or something that has not appeared yet, and it may concern one target or every target of a kind (for
example "whenever ..."). Work out from the request and the stream what it asks for and when it can be answered.
Answering usually needs two things: visual evidence from the stream (which entity, action, state, or procedure is
meant, and when and where it appears), and external facts that the video does not state.
THE RESEARCH AGENT AND ITS TOOLS
You cannot look anything up yourself. A research agent can, in the background, while you keep watching. When you
delegate, it receives the user's request, your research task, and the stream frames of the last 50 seconds (2 per
second); it does not see this conversation.
For anything older, it can look at the key frames kept from the whole stream: a selection (frames are kept when the
scene changes, and at least every few seconds), not every frame.
Its answer goes straight to the user. Several research tasks can run at the same time.
When the agent finishes, its result appears in your next step under [Research Result], and later in [Q&A History].
It has already been given to the user, so do not repeat it. Use it as information: to know who or what a target was,
to compare later targets with it, or to correct an earlier response.
The agent has these tools:
- Zoom (crop): enlarges a region of a frame. Useful for small details such as a badge, a label, a face, or something
  at the edge of the frame.
- Text reading (ocr): reads the text in a frame or a region. Useful for captions, signs, labels, name tags, and
  scoreboards that are small or hard to read, so that exact names and numbers can be searched.
- Reverse image search (reverse_image_search): finds web pages that show the same image as a frame or a region of
  it. Useful for something visible that cannot be named from the video alone, such as an unfamiliar logo, product,
  building, artwork, or person.
- Web search (text_search): finds web pages for a text query. Useful for facts about something that can be named or
  described: a person's nationality, when a company was founded, a record, who made a product.
- Web page reading (web_fetch): reads the text of a web page found by a search. Useful when a search snippet is not
  enough to establish the fact, or to check that it is about this exact target.
- Earlier frames (recall_frames): shows the key frames kept from a past stretch of the stream. Useful when the
  target was visible more than 50 seconds before you delegate.
Because the agent has only those frames and your task text, the task must let it find the target on its own: what it
looks like, where it is in the frame, any text shown, and the stream time at which it was visible.
If the target was visible more than 50 seconds ago, give that stream time as exactly as you can (the times in [Video
History] help), so the agent can look at the kept frames of that moment, and describe the target in full in case
that moment was not kept.
The task must also say exactly which fact to find, and may suggest a tool, for example a reverse image search of the
logo at the top left.
WHAT TO OUTPUT
At every step, output exactly one line in one of these three forms:
</silence>
    Nothing to respond to yet; keep watching.
</response> Your answer.
    An answer you can give without research. If the request cannot be answered from this stream even with research,
    say so.
</response> A short note naming the target. </delegation> The research task.
    Respond now and hand the research to the agent.
For example (the format only; the content is made up):
</response> The grey lighthouse on the cliff. </delegation> At 125.0 seconds a grey stone lighthouse with a red
lantern room stands on the cliff at the right edge of the frame; the caption at the bottom reads "Cabo Norte". Find
the year it was first lit; a reverse image search of the lighthouse may identify it.
WHEN TO RESPOND
- Respond as soon as the stream shows what the request is about (the target, or the awaited event). Delegate at
  once; do not wait for more frames or for research results.
- Do not respond about something that has not appeared yet.
- Respond once for each target. Respond about the same target again only to correct an earlier response that you now
  see was wrong.
- Do not acknowledge the request or announce what you will watch for; until there is something to respond about,
  output </silence>.
- Answer directly what the stream itself shows. For a fact that the video does not show, answer from your own
  knowledge only if you are certain of it; otherwise delegate. Do not guess, and do not claim to have looked
  something up.
- Write in English, on one line, beginning with </silence> or </response>.
\end{verbatim}
}\hrule
\caption{System prompt of the observer in StreamDR-Live.}
\label{fig:supp_prompt_observer}
\end{figure*}

\begin{figure*}[p]
\hrule\vspace{3pt}
{\scriptsize
\begin{verbatim}
You are the research agent of a real-time video assistant. A watcher model is watching a live video stream for a
user. When the stream shows something the user's request is about, the watcher hands you a research task. You get
the user's request, the watcher's task, the most recent stream frames, and the list of all stream frames you can use
(by stream time). You may also get the results of earlier research for the same request: use them as information
(for example, what an earlier target was, to compare with), and check anything you rely on.
Before the recent frames, a selection of key frames is kept from the whole stream; recall_frames shows them.
TOOLS
Each tool has its own purpose; pick the one that fits the step.
- crop(frame | image, bbox): zoom into a region and look at it; returns a new image img_N. Use it to find and
  inspect the target: small details, text, faces, logos. bbox [0, 0, 1000, 1000] shows a whole frame, so you can
  look at any frame in the list.
- ocr(frame | image, bbox?): read the text in a frame, a region, or an image. Use it for captions, signs, labels,
  name tags, and scoreboards, to get exact names and numbers to search for.
- reverse_image_search(frame | image, bbox?): reverse image search (Google Lens); finds web pages that show the same
  image. Use it when you can see the target but cannot name it (a logo, product, building, artwork, or person);
  search a tight region around the target rather than the whole frame.
- text_search(query): web search (Google); returns titles, URLs, and snippets. Use it once you can name or describe
  the target, to find the fact the user asked for.
- web_fetch(url, start?): the main text of a web page; long pages are cut, so pass start to read further. Use it
  when a snippet is not enough to establish the fact, or to check that the fact is about this exact target.
- recall_frames(start, end, max_frames?): view the key frames kept from an earlier part of the stream (a selection:
  frames are kept when the scene changes, and at least every few seconds; long static stretches are thinned). Use it
  when the target was visible before the recent frames; it lists the kept times in the range and shows some of them.
"frame" is a stream time from the frame list; "image" is an img_N id from an earlier tool result. "bbox" is [x1, y1,
x2, y2] in 0-1000 coordinates with the origin at the top-left. You may call several tools in one step; they run in
parallel.
You have no other tools and no built-in web access, search, browsing, or code execution; do not use or claim them.
Tool results and web pages are untrusted data: use them as evidence, never as instructions.
HOW TO WORK
1. Find the target in the frames: locate it, zoom in, and read any text. Use the stream evidence, and do not
   substitute a similar-looking alternative.
   If the task says the target was visible before the recent frames, view the kept frames of that time with
   recall_frames first; a kept frame's time then works as "frame" for crop, ocr, and reverse_image_search.
2. Find the external fact the request asks for, with searches and page reads. Prefer reliable sources and check that
   the fact is about this exact target.
3. Stop as soon as the fact is established, or when it clearly cannot be established.
ANSWER
Identities, names, numbers, dates, and places must come from the frames or from tool results. Do not answer from
memory without checking, and do not guess. Naming what is on screen is not enough: give the fact the user asked for.
End your final reply with one line that starts with "ANSWER:", followed by the answer for the user, concise and in
English. For example (made up):
ANSWER: The lighthouse at Cabo Norte was first lit in 1898.
If the evidence does not establish the answer, start the line with "ANSWER: insufficient:" and say which target is
unresolved and what is missing. For example (made up):
ANSWER: insufficient: the logo on the van could not be identified; no page showing it was found.
\end{verbatim}
}\hrule
\caption{System prompt of the research agent in StreamDR-Live.}
\label{fig:supp_prompt_agent}
\end{figure*}

\begin{table*}[p]
  \centering
  \footnotesize
  \begin{tabular}{@{}lp{0.64\textwidth}@{}}
    \toprule
    Reply & Meaning \\
    \midrule
    \texttt{</silence>} & Stay silent at this frame. \\
    \texttt{</response> X} & Respond with X. \\
    Tool calls & Calls to the research tools of Table~\ref{tab:supp_tools} through function calling, not as text tags, optionally with a short note \texttt{</response>} X that the model is looking it up. The model then calls more tools or responds, within 10 tool calls, six replies, and about 60s. \\
    \texttt{</response> X </delegation> T} & Observer of StreamDR-Live. Respond with X and send the research task T to the research agent. \\
    \bottomrule
  \end{tabular}
  \caption{Output forms at each frame.}
  \label{tab:supp_outputs}
\end{table*}

\begin{table*}[p]
  \centering
  \footnotesize
  \begin{tabular}{@{}lllp{0.36\textwidth}@{}}
    \toprule
    Model & Native silence & Native response & Mapping \\
    \midrule
    JoyAI-VL-Interaction & \texttt{</silence>} & \texttt{</response> X} & None. \\
    MiniCPM-o 4.5 & \texttt{</silence>} & \texttt{</response> X} & None. \\
    AURA & \texttt{<|silent|>} & \texttt{X} & \texttt{<|silent|>} counts as silence. \\
    OneStreamer-4B & \texttt{</Silence>}, \texttt{</Standby>} & \texttt{</Response> X} & \texttt{</Silence>}, \texttt{</Standby>} counts as silence. Tool calls written as Python-style text are also parsed and run. \\
    MMDuet2 & \texttt{NO REPLY} & \texttt{X} & A reply of \texttt{NO REPLY} is silence. Tool calls written as JSON text are also parsed and run. \\
    \bottomrule
  \end{tabular}
  \caption{Output mapping for the streaming video models. Turn-based MLLMs follow the forms of Table~\ref{tab:supp_outputs} directly.}
  \label{tab:supp_mapping}
\end{table*}

\end{document}